\documentclass{article}

\usepackage{arxiv}

\usepackage[utf8]{inputenc} 
\usepackage[T1]{fontenc}    
\usepackage{hyperref}       
\usepackage{url}            
\usepackage{booktabs}       
\usepackage{amsfonts}       
\usepackage{nicefrac}       
\usepackage{microtype}      
\usepackage{lipsum}		
\usepackage{graphicx}
\usepackage{natbib}
\usepackage{doi}
\usepackage{caption}
\usepackage{subcaption}

\title{MalFake: A Multimodal Fake News Identification for Malayalam using Recurrent Neural Networks and VGG-16}

\author{Adhish S. Sujan\\
	School of Digital Sciences\\
	Kerala University of Digital Sciences-\\Innovation and Technology\\
	Thiruvananthapuram, India \\
	\texttt{adhish.ds22@duk.ac.in} \\
        \And
        Ajitha V.\\
	School of Digital Sciences\\
	Kerala University of Digital Sciences-\\Innovation and Technology\\
	Thiruvananthapuram, India \\
	\texttt{ajitha.ds22@duk.ac.in} \\
	\And
        Aleena Benny\\
	School of Digital Sciences\\
	Kerala University of Digital Sciences-\\Innovation and Technology\\
	Thiruvananthapuram, India \\
	\texttt{aleena.ds22@duk.ac.in} \\
        \And
        Amiya M P\\
	School of Digital Sciences\\
	Kerala University of Digital Sciences-\\Innovation and Technology\\
	Thiruvananthapuram, India \\
	\texttt{amiya.ds22@duk.ac.in} \\
	\And
	{V. S. Anoop} \\
        Applied NLP Research Group\\
	School of Digital Sciences\\
	Kerala University of Digital Sciences-\\Innovation and Technology\\
	Thiruvananthapuram, India \\
	\texttt{anoop.vs@duk.ac.in} \\
}

\date{}

\renewcommand{\undertitle}{}

\hypersetup{
}

\begin{document}
\maketitle

\begin{abstract}
The amount of news being consumed online has substantially expanded in recent years. Fake news has become increasingly common, especially in regional languages like Malayalam, due to the rapid publication and lack of editorial standards on some online sites. Fake news may have a terrible effect on society, causing people to make bad judgments, lose faith in authorities, and even engage in violent behavior. When we take into the context of India, there are many regional languages, and fake news is spreading in every language. Therefore, providing efficient techniques for identifying false information in regional tongues is crucial. Until now, little to no work has been done in Malayalam, extracting features from multiple modalities to classify fake news. Multimodal approaches are more accurate in detecting fake news, as features from multiple modalities are extracted to build the deep learning classification model. As far as we know, this is the first piece of work in Malayalam that uses multimodal deep learning to tackle false information. Models trained with more than one modality typically outperform models taught with only one modality. Our study in the Malayalam language utilizing multimodal deep learning is a significant step toward more effective misinformation detection and mitigation.
\end{abstract}


\section{Introduction}
The spread of false or misleading information as if it were true news has become a big concern in our current digital age. This issue is not confined to major languages alone; it also impacts regional languages like Malayalam. During the COVID-19 pandemic, a plethora of false information circulated online. They even said that treatments with vinegar, tea, and salt water could be used as effective remedies\cite{satu2021tclustvid}\cite{varma2021systematic}. Identifying fake news before it spreads is crucial to prevent it from spreading and any associated potential damage it causes. In the past, many fake news identification approaches have been reported in the machine learning literature with varying degrees of success \cite{chen2023using}\cite{kondamudi2023comprehensive}\cite{jain2023confake}. With the advancement in computing and social media proliferation, new advanced approaches such as multi-modal news content started being shared by people across the globe \cite{varghese2022deep}\cite{anoop2023public}\cite{lekshmi2022sentiment}. The text-only approaches may not give good detection accuracies in this case, leading to the development of multi-modal fake news detection techniques\cite{comito2023multimodal}\cite{jing2023multimodal}. The knowledge graphs-based approaches have also become prominent in identifying fake news\cite{pan2018content}. Multi-modal fake news identification itself is very challenging, and detecting fake news from low-resource languages is even more challenging\cite{jain2023confake}\cite{praseed2023hindi}.

This study presents a novel approach to detect and counteract misinformation in Malayalam, which is a low-resource language that is highly agglutinative as well\cite{akhil2020parts}\cite{rajimol2020framework}.  Malayalam is a South Indian language with a rich cultural heritage and a growing online presence. With a substantial Malayalam-speaking population, it is crucial to protect the integrity of information within this linguistic community. Fake news specifically targets regional languages like Malayalam, capitalizing on linguistic and cultural nuances often overlooked in the fight against disinformation. We aim to achieve this by incorporating diverse data types, such as text and images, to safeguard the accuracy of information within the Malayalam-speaking community. As fake news increasingly targets regional languages, there is a growing need for specialized solutions. This paper aims to tackle these issues directly and ensure the responsible sharing of accurate news and information in the digital age. Fake news has become a pervasive challenge in our digital landscape, where misinformation can easily spread across various online platforms, including social media. The consequences of fake news are far-reaching, as it can incite panic, manipulate public opinion, and erode trust in credible sources of information.

\indent The proposed methodology includes natural language processing and deep learning techniques \cite{alonso2021multimodal}. NLP algorithms analyze the text content of news articles, social media posts, and other sources. In parallel, deep learning techniques are applied to analyze images associated with news stories. This includes using deep neural networks to detect manipulated or doctored images and identify misleading visual content. This combination of NLP and deep learning techniques allows a more comprehensive assessment of the presented information. The effectiveness of any fake news detection model relies on the quality and diversity of the data sources used for training and testing. In the case of Malayalam, a robust dataset of news articles, social media posts, and multimedia content in the language is essential. Collaborations with local news outlets, social media platforms, and community-driven initiatives provide access to a broad range of Malayalam content. Developing a fake news detection model for Malayalam involves creating a hybrid system that combines NLP and deep learning components. Using supervised learning methods, the model is trained on a labeled dataset of real and false news in Malayalam. Convolutional neural networks and recurrent neural networks, two deep learning techniques, create a model that distinguishes between the two groups. The major contributions of this paper are outlined as follows:
\begin{itemize}
    \item Surveys some of the recent state-of-the-art approaches in multi-modal fake news identification approaches.
    \item Proposed a multi-modal framework incorporating modules for understanding images and text to identify fake news.
    \item Experimentally verifies the usefulness of the proposed approach in terms of precision, recall, and accuracy measures.
\end{itemize}
\section{Related studies}
Some of the most innovative methods for multi-modal false news identification documented in the machine learning literature are shown in the paper of Segura et al. \cite{segura2022multimodal}. The paper describes a method for detecting fake news using unimodal and multimodal approaches. This paper examines the advantages of combining textual and visual data and assesses fake news using the Fakeddit dataset. The authors extract information from images using pre-trained models, which they then mix with textual features to create a multimodal representation of each instance. They discovered that, with an accuracy of 87\%, the multimodal approach based on CNN that took into account both text and image data yielded the best results. An approach for online social networks (OSNs) using text and visual data is presented by  Santosh Kumar Uppada et al. \cite{uppada2022image}. They emphasize how misleading visual news may be and how it affects people psychologically. To analyze images and text, the authors use various methods, such as Error Level Analysis, VGG-16, VGG-19, Xception, Inception-Resnet50, and BERT models. The research employs approximately 1 million samples from the Fakeddit dataset, which includes text, photos, metadata, and comments. The technique, including the framework, picture editing, polarity-based fake post-detection models, and fusion models, is covered in detail in the paper. Additionally, it explains the outcomes of the experiments and offers an error analysis of the suggested model.

Anit Sara Santhosh et al. published a paper to detect Fake News using Machine Learning \cite{bijimol2022malayalam}. The goal of the research is to develop a machine-learning model that can reliably identify if news in Malayalam is true or not. By utilizing NLP strategies and machine learning methods like the TF-IDF Vectorizer and Passive Aggressive Classifier, the authors hope to accomplish the problem. The study's approach entails gathering data, preprocessing it, creating a model, analyzing it, and reporting the findings. The authors used a dataset with 316 rows and 2 columns that had labels that were both fake and true and headings that indicated whether a piece of Malayalam news was true or fake. The work of Sudhanshu Kumar et al. \cite{kumar2022fake} discusses a strategy for handling fake news in Hindi. After employing NLP techniques for feature engineering and pre-processing, the proposed method uses machine learning and deep learning to classify news articles. The study uses two independent datasets: FNC-1 and a news dataset in Hindi. The methods for feature extraction employed include TF-IDF and bag-of-words. Multilayer Perceptron, Multinomial Naive Bayes, Support Vector Machine, Logistic Regression, and Long Short-Term Memory (LSTM) are some of the categorization techniques employed.

A detailed description of the preprocessing, feature extraction, classification, and prediction algorithms can be found in the Priyanshi Shah et al\cite{shah2020multimodal} publication. This article outlines a thorough plan to stop the spread of false information on social media. It initiates with a Textual Feature Extractor, utilizing sentiment analysis to gauge the emotional tone of news articles, a critical step as fake news often manipulates emotions to deceive readers. In parallel, a Visual Feature Extractor processes accompanying images, involving resizing, grayscale conversion, and advanced techniques like K-means clustering and Discrete Wavelet Transform (DWT) to extract significant visual features. These extracted features serve as crucial inputs for subsequent steps. The optimized feature vectors from textual and visual components are then fine-tuned using a Cultural Algorithm, which expertly combines normative and situational knowledge to refine the feature set while minimizing computational costs. Finally, the Fake News Detector merges the optimized feature vectors and employs a kernel Support Vector Machine (SVM) for classification, leveraging the power of SVMs to distinguish between real and fake news. The classifier undergoes rigorous training with labeled data, aided by cross-validation to ensure its robustness. This holistic approach, integrating both textual and visual information, is designed to significantly enhance fake news detection accuracy, as substantiated through a series of extensive experiments on real-world datasets, where it consistently outperforms existing methods by an average margin of 9\% in terms of accuracy.

\indent In the work of Suryavardan et al. \cite{suryavardan2023factify}, a multimodal fact-checking dataset called FACTIFY 2 is introduced as a solution to the rising problem of fake news and disinformation on the internet. By adding additional data sources and satirical articles, this dataset expands upon its predecessor, FACTIFY 1, and now contains 50,000 new data instances. Based on the inclusion of textual and visual data between claims and their supporting documents, FACTIFY 2 encompasses five separate categories: Support Text, Support Multimodal, Insufficient Text, Insufficient Multimodal, and Refute. We offer a baseline model based on BERT and Vision Transformer (ViT), showcasing its efficacy with a test set F1 score of 65\%. An approach for SpotFake fake news proliferation on social media is addressed in the paper of Singhal et al. \cite{singhal2019spotfake}. The authors introduce SpotFake, a novel multi-modal framework for detecting fake news without additional subtasks. This framework leverages textual and visual features, utilizing BERT for text analysis and pre-trained VGG19 for image features. The experiments conducted on Twitter and Weibo datasets demonstrate that SpotFake outperforms current state-of-the-art models significantly, improving accuracy and precision for fake news detection. The paper also highlights the importance of using multiple modalities for improved fake news detection, as supported by a public survey that revealed the benefits of combining text and image information. Overall, SpotFake represents a promising approach for addressing the challenge of fake news detection in the age of multimedia information.

\indent In a paper by Rina Kumari et al. \cite{kumari2021amfb}. (2021), they discuss detecting news using a multimodal factorized bilinear pooling approach. Fake news is purposely created to mislead people. They can have consequences for society. With abundant multimedia information on platforms like Twitter, Facebook, and blogs, it becomes challenging to identify news. To address this issue, the authors proposed a framework that combines visual information to maximize their correlation. By analyzing posts at a stage in the network, the model determines whether they are genuine or fake. The proposed framework surpasses existing models by 10 percentage points, achieving better performance with balanced F1 scores for both fake and real classes. The framework consists of four sub-modules; Attention Based Stacked Bidirectional Long Short Term Memory (LSTM), Attention Based Multilevel Convolutional Neural Network (CNN) ABM CNN RNN, MFB, and Multi-Layer Perceptron (MLP). Importantly this approach does not require any user or network details. The effectiveness of this method is evaluated on two datasets; Twitter and Weibo. Additionally, the model's complexity is significantly reduced compared to state-of-the-art models.

Alessandro Bondielli et al.'s paper for the MULTI-false-DetectiVE challenge for the EVALITA 2023 campaign \cite{bondielli2023multi} examines multimodality in the context of false news. Gaining an understanding of the interaction between text and visuals and assessing the efficacy of multimodal false news detection systems are among the responsibilities. The study makes the case that the issue is unresolved and suggests some potential paths forward. In recent years, disinformation has developed into a potent strategic tool, particularly regarding actual occurrences covered as breaking news. The initial "Infodemic" that followed the COVID-19 epidemic in recent years has shown the distorted usage of online social media. Large language models and other related approaches are also covered in the article. Pengwei Zhan et al.'s study \cite{chen2022cross} describes a strategy for stopping the severe spread of fraudulent information on social media. There are serious consequences due to fake news spreading on social media. Automatically identifying fake news is essential to reducing these effects. To more effectively combine textual and visual information for false news identification, Multimodal Co-Attention Networks (MCAN) are suggested. MCAN outperforms state-of-the-art techniques and can learn inter-dependencies among multimodal characteristics. It is more effective to detect fake news when text and images are combined, according to recent research. Nevertheless, conventional expert identification techniques disregard pictures' semantic properties.

\indent In the paper \cite{silva2021embracing}, researchers from the University of Melbourne provide a novel framework for cross-domain detection of false news using multi-modal data. The issue of fake news is a serious social concern that cannot be solved quickly via manual research. To detect false news, most research investigates supervised training models using various news record modalities; nevertheless, the effectiveness of these methods often decreases when news records originate from diverse domains. The scientists suggest an unsupervised method for choosing meaningful unlabeled news records for manual labeling to increase the labeled dataset's domain coverage. When the suggested fake news model and selective annotation method are combined, crossing-domain news datasets may perform at cutting-edge levels, and typically appearing domains in news datasets can see significant gains in performance. According to the authors, automatic fake news identification has emerged as a major issue that is getting much attention from researchers. However, it is challenging to spot false news since existing methods are restricted to a specific industry, such as politics, entertainment, or healthcare. The study \cite{wu2021multimodal} describes the unique Multimodal Co-Attention Networks (MCAN) presented by the Association for Computational Linguistics (ACL-IJCNLP) to better merge textual and visual features for false news identification. Fake news has become increasingly prevalent thanks to social media, which has had detrimental effects. Recently, tweets containing photos have gained popularity on social media because they provide richer content and draw in more people than tweets that solely contain text. This benefit is also fully used by fake news to tempt and confuse readers. MCAN outperforms state-of-the-art techniques and can learn inter-dependencies among multimodal characteristics. It is more effective to detect false news when text and the linked image are combined, according to recent research. Nevertheless, current methods for feature fusion and extraction are not sufficiently fine-grained.

\indent The proliferation of fake news on social media is a serious problem with documented negative impacts on individuals and organizations described in the paper of \cite{wu2021multimodal}. Researchers are building detection algorithms with an aim for high accuracy. This paper presents a novel approach using a Cultural Algorithm with situational and normative knowledge to detect fake news using text and images. The proposed method outperforms the state-of-the-art methods for identifying fake news in terms of accuracy by 9\% on average. Fake news is intentionally written to confuse viewers, making it nontrivial to identify simply based on news content. The article discusses the prevalence of fake news and the need for research on automating the detection of false information and verifying its accuracy, which is discussed in the paper of \cite{suryavardan2023findings} . It presents the outcome of the Factify 2 shared task, which provides a multi-modal fact verification and satire news dataset, as part of the DeFactify 2 workshop at AAAI 2023. The data calls for a comparison-based approach to the task by pairing social media claims with supporting documents, with both text and image, divided into 5 classes. The best performances came from using DeBERTa for text and Swinv2 and CLIP for images. The highest F1 score averaged for all five classes was 81.82\%. The article highlights the difficulty in uncovering misleading statements before they cause significant harm and the scarcity of available training data hindered automated fact-checking efforts. The rapid distribution of news across numerous media sources, particularly on social media, has led to the fast development of erroneous and fake content.\\

The emergence of multi-modal fake news has caused social harm, making it difficult to detect it accurately.A study on this is done by \cite{peng2022effective}. Traditional detection methods involve fusing different modalities without considering their impacts, leading to low accuracy. To address this, a new attention and adversarial fusion method, based on the pre-training language model BERT, has been developed. The attention mechanism captures differences in modalities, while the adversarial mechanism captures correlations between them. The proposed new method achieves 5\% higher accuracy than the traditional method. To counter the proliferation of fake news, there is a need for in-depth research into automatic monitoring methods for fake news. The article discusses the main problem in detecting fake news and how to distinguish it according to its characteristics, including sources, texts, and attached pictures. It also highlights the importance of considering the complementarity of different features to improve detection accuracy.

\section{Materials and Methods}
\subsection{Long-Short Term Memory (LSTM)}
LSTM is a specific kind of recurrent neural network (RNN) that was created to solve the intrinsic drawbacks of conventional RNN, including the disappearing gradient issue.\cite{giachanou2020multimodal} The LSTM design is well-known for solving this issue and enabling it to identify both short and long-term relationships between consecutive data efficiently. Because of this property, LSTM is well-suited to tasks that need a thorough comprehension of context, the recognition of complex patterns encompassing extended periods, and the discovery of remote links between parts. The main component of the multimodal fake news detection technique is Long Short-Term Memory (LSTM), which has extensive training in interpreting textual data. Its main job is to discover short- and long-term relationships within sequences to identify linguistic signs typical of fake news in Malayalam. Text from diverse sources is methodically processed by LSTM, which carefully looks for language patterns and irregularities. This advanced method improves accuracy by identifying minute language details associated with false news.
\subsubsection{LSTM architecture}
\begin{figure}
	\centering
	\includegraphics[width=\textwidth,height=7cm]{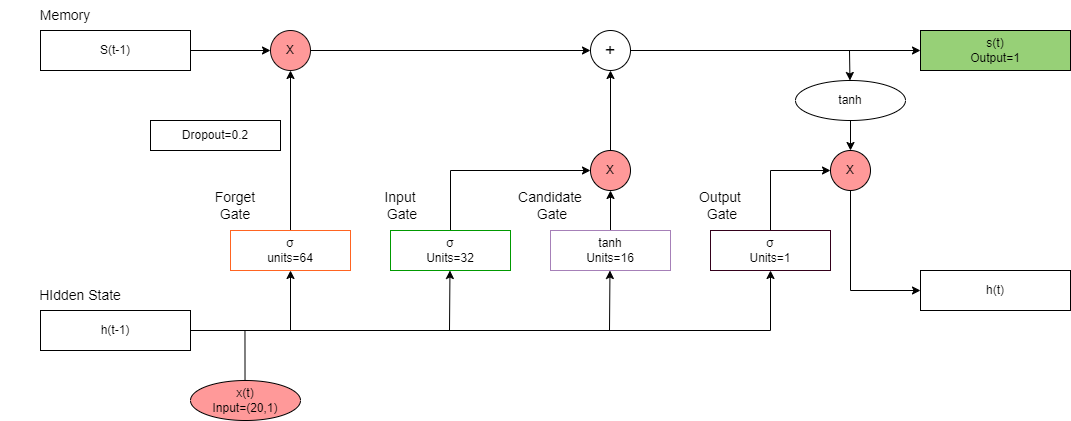}
	\caption{Architecture of a Long Short-term Memory}
	\label{fig:fig2}
    \end{figure}

A specific type of recurrent neural network (RNN) called the LSTM was designed to be developed to tackle the issue with standard RNNs' vanishing gradient. Three different types of gating units—the input gate, forget gate, and output gate—control the information flow into and out of the memory cell, which is the central component of the architecture. The memory cell itself is capable of maintaining its state over time. Information entering a memory cell is managed by the input gate, and information leaving the memory cell is managed by the forget gate. The output gate manages the flow of information from the memory cell to the network's output. A sigmoid activation function, which produces a value, is used to build each gate. A sigmoid activation function, which outputs a number between 0 and 1, is used to build each gate. This function establishes the quantity of information that can pass through the gate. The activations of the gates and the memory cell are calculated using a set of weights and biases included in the LSTM architecture in addition to the gating units. Backpropagation through time (BPTT), a variation of the backpropagation algorithm used to train RNNs, is the method used to learn these weights and biases during training. The LSTM design has advanced the best available for numerous demanding situations and has been demonstrated to be successful at capturing long-term temporal dependencies. The architecture of the LSTM model is shown in Figure \ref{fig:fig2}.
We train our LSTM model with pre-trained word embeddings in our Multimodal Fake News Detection project, giving it a solid linguistic base. Because of these embeddings, our model can learn semantic understanding and recognize complex linguistic cues in Malayalam. These pre-trained embeddings are important because they simplify training and improve our model's ability to spot minute patterns and irregularities in text. Pre-trained embeddings serve as a link between the analytical capabilities of our LSTM model and the linguistic complexity of Malayalam, improving the accuracy and precision of our fake news detection system.
\subsection{VGG-16}
VGG-16 stands for "Visual Geometry Group 16. It is a versatile deep learning architecture primarily used for image classification, object recognition, feature extraction, transfer learning, and benchmarking in various computer vision applications. Its deep structure and effectiveness on standard datasets have made it popular in the deep learning community. The network's depth, denoted by "16" in VGG-16, comprises 16 weight layers. VGG-16's 13 convolutional layers, which are in charge of extracting complex patterns and information from input images, are its fundamental component. The fact that these convolutional layers continually use 3x3 convolutional filters, scan the input with a stride of 1, and use "same" padding to preserve the spatial dimensions is noteworthy. This makes sure that crucial spatial information is preserved across the network. VGG-16 adds 2x2 max-pooling layers with a stride of 2 after each block of convolutional layers.\cite{tammina2019transfer} The feature maps are downsampled using these layers, which gradually reduce the feature maps' spatial size while enhancing their depth. This architectural decision helps the network efficiently capture hierarchical features. The architecture of VGG-16 model is shown in Figure\ref{fig:fig3}.
    \begin{figure}
        \centering
        \includegraphics[width=10cm,height=7cm]{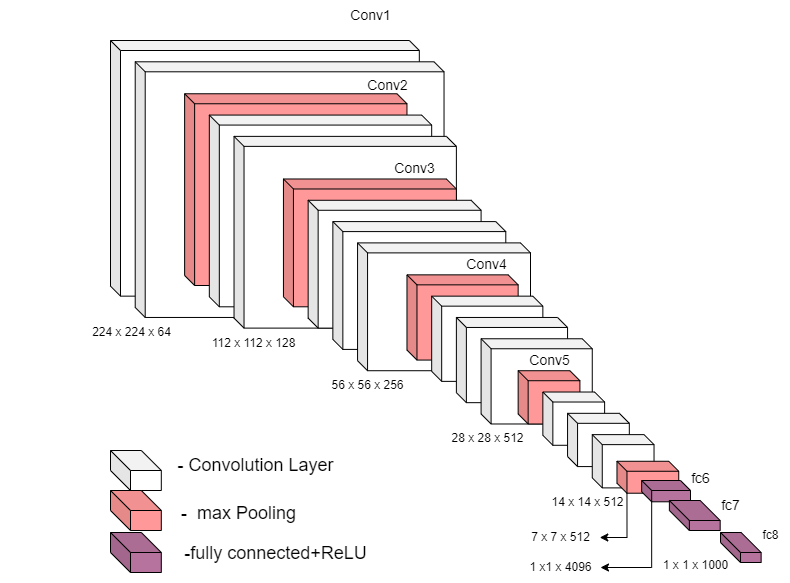}
        \caption{VGG 16 Architecture.}
        \label{fig:fig3}
    \end{figure}
\subsection{Dataset}
The Multimodal Fake News Detection dataset has been thoughtfully assembled to support the development and evaluation of algorithms specifically tailored for the identification of fake news articles across a range of online platforms, including Facebook and various news websites such as Manoramaonline, One India, Anweshanam, Mangalam, Janam TV, 24 News, Asianet News, Samayam, and VishvasNews. This dataset comprises four fundamental components: \textit{news\_headline}, \textit{news\_url}, \textit{image\_url}, and \textit{image\_name}, alongside manually assigned binary labels, designating the veracity of each news article, with 0 signifying fake and 1 indicating true.
\begin{itemize}
\item \textbf{News\_headline}: This component encompasses the headline or title of each news article, serving as the textual basis for subsequent analysis and classification.
\item \textbf{News\_url:} It denotes the URL of the source from which the news article originates. This information proves invaluable for cross-referencing and source validation.
\item \textbf{Image\_url:} Corresponding to the URL of the associated image for each news article, images serve to provide supplementary context and vital visual cues to detect fake news.
\item \textbf{Image\_name:} An automatically generated identifier for the associated image, facilitating the seamless connection between the image and its respective news article.
\item \textbf{Target Class (Label):} Each news article within the dataset is categorized as either "fake" (0) or "true" (1) based on a meticulous process of manual assessment and content verification.
\end{itemize}
\begin{figure}
        \centering
        \includegraphics[width=12cm,height=7cm]{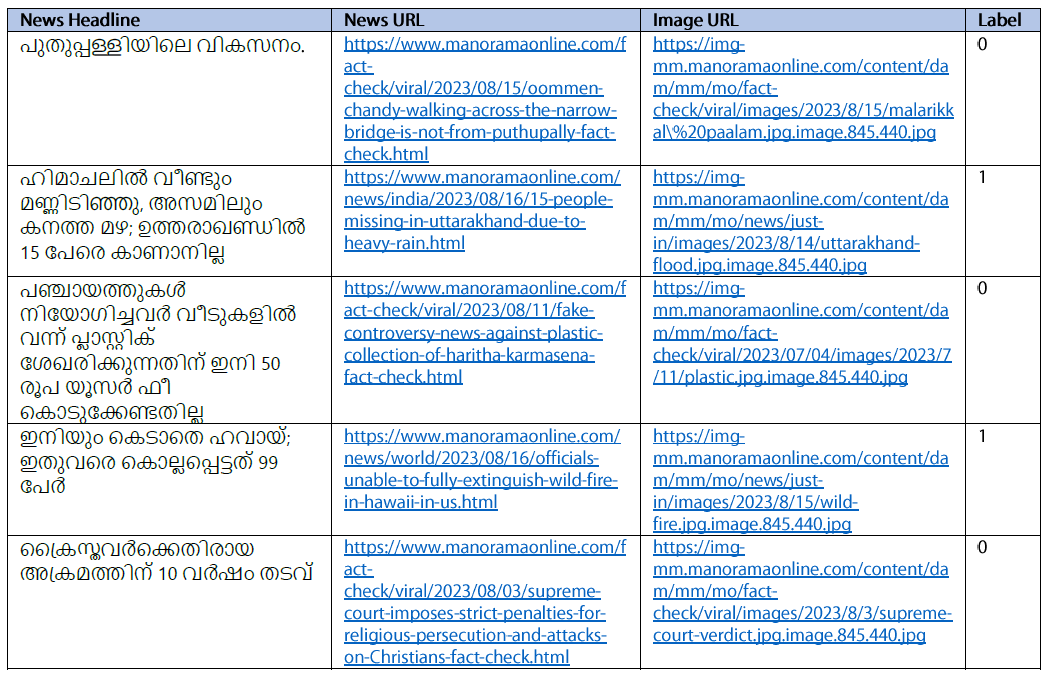}
        \caption{A snapshot of the dataset used in this work}
        \label{fig:fig3}
    \end{figure}
The dataset encompasses 1852 data points, including 926 fake news articles, and there are 926 instances of true news. Data acquisition involved compiling data from various sources, including Facebook and multiple news websites. For each news article, critical information, including \textit{news\_headline}, \textit{news\_url}, and \textit{image\_url} was meticulously extracted and integrated into the dataset. The \textit{image\_name} was algorithmically generated from the \textit{image\_url} to ensure coherent association. Determining labels for fake news (0) and true news (1) was conducted with scrupulous attention to manual content verification. A snapshot of the dataset is shown in Figure 3.
\section{Proposed Approach}
\begin{figure}
	\centering
	\includegraphics[width=\textwidth]{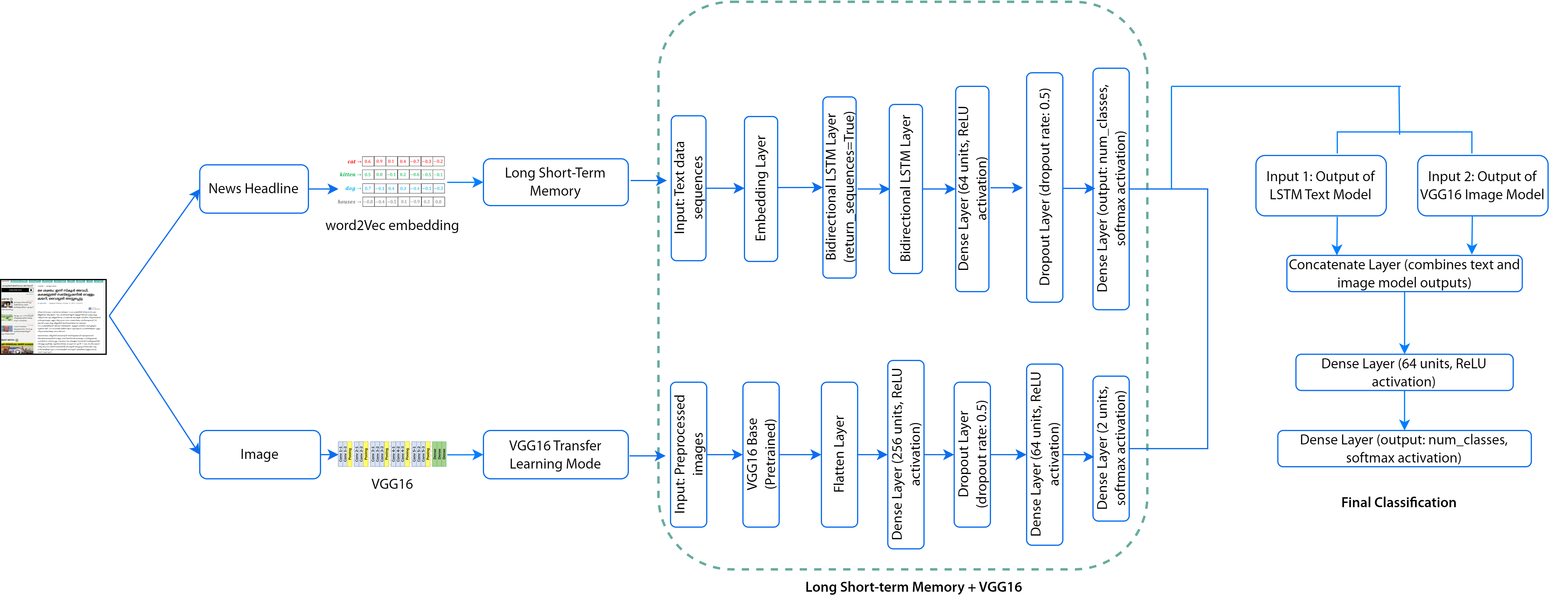}
	\caption{Our proposed model for Multimodal Fake News Detection.}
	\label{fig:fig4}
    \end{figure}

This section outlines our methodology for detecting fake news in the Malayalam language through a multimodal framework. Our methodology integrates text and image information, harnessing the power of natural language processing and deep learning techniques to enhance the accuracy and robustness of our model. The overall workflow and architecture for our proposed approach are shown in Figure \ref{fig:fig4}.
\subsection{Feature Extraction}
\subsubsection{Text Feature Extraction}
Our approach to identifying fake news relies heavily on textual content as a data source. To ensure that the textual data is translated into a format that can be analyzed, several essential steps must be taken during the text feature extraction process. As our primary text data, we first collect headlines from news items. These headlines contain important information that may indicate whether it is reliable. We first preprocess the Malayalam text before moving on to feature extraction. Special characters, HTML tags, digits, and other components that are not necessary for the task at hand are removed in this stage. To generalize numerical data, we swap out numeric values with the generic term "NUM." This preparation enables a clearer and more reliable representation of the text. Next, we tokenize the cleaned headlines into words, following Malayalam language conventions. Tokenization breaks the text into individual units, allowing us to work with words as separate entities. our approach involves the utilization of Word2Vec for textual feature extraction, allowing us to extract the semantic associations between words in the Malayalam text and build word embeddings. These word vectors play a crucial role in enhancing the understanding of textual content and improving the accuracy of our fake news detection system.
\subsubsection{Image Feature Extraction}
Images included with news items often provide insightful visual information that might enhance the textual content. We use the VGG16 transfer learning model to extract useful image characteristics to make use of this. This model can recognize various visual patterns and features because it was pre-trained on a sizable image dataset. The image feature extraction involves loading and preprocessing the images from the dataset. Images are resized to match the input size of the VGG16 model and converted to the appropriate format. After preprocessing, we utilize the VGG16 model to obtain the last layer's output, representing the most crucial image features. To ensure consistency with the text data, we flatten the extracted image features and pad them if necessary to match the maximum sequence length used for text data. This alignment is crucial for subsequent fusion and analysis.
\subsection{Classifier and Model Selection}
\subsubsection{LSTM-Based Text Model}
The core of our text-based fake news detection lies in an LSTM-based model, carefully designed to process the textual content effectively. The architecture of our LSTM Text Model consists of the following layers:
\begin{itemize}
    \item Embedding Layer: This layer converts textual data into numerical vectors, enabling the model to understand and process the text effectively. It transforms words into dense vectors where similar words have similar representations.
    \item Bidirectional LSTM Layer (return sequences=True): The 'return sequences' parameter is set to True in this layer, which uses Bidirectional Long Short-Term Memory (LSTM) units. With the help of bidirectional LSTMs, which record contextual information in both the forward and backward directions, the model can comprehend the context of the text in greater detail.
    \item Bidirectional LSTM Layer: Similar to the layer above, this Bidirectional LSTM layer further enhances the text's contextual understanding.
    \item Dense Layer (64 units, ReLU activation): This dense layer introduces non-linearity to the model by applying Rectified Linear Unit (ReLU) activation. It consists of 64 units, which allows the model to figure out detailed relationships within the information.
    \item Dropout Layer (dropout rate: 0.5): During training, the Dropout Layer randomly sets a portion of the input units to zero to prevent overfitting. In our model, the dropout rate is set to 0.5.
    \item Dense Layer(output: num classes, softmax activation): In the context of multi-class classification, the number of units in the final Dense Layer of the neural network architecture is set to match the number of classes ('num classes'). To derive class probabilities, the softmax activation function is applied.
\end{itemize}
\subsubsection{VGG16 Transfer Learning Model}
For image-based fake news detection, we employ a VGG16-based model, leveraging the power of transfer learning. The architecture of our Image Model includes the following layers:
\begin{itemize}
    \item Flatten Layer: This layer is essential to reshape the output of the VGG16 base into a one-dimensional vector, preparing it for further processing.
    \item Dense Layer (256 units, ReLU activation): The Dense Layer with 256 units captures high-level image features, allowing the model to recognize complex visual patterns.
    \item Dropout Layer (dropout rate: 0.5): This layer has a dropout rate of 0.5 which is applied to prevent overfitting and maintain model generalization.
    \item Dense Layer (64 units, ReLU activation): This additional Dense Layer with 64 units captures more specific image details and patterns.
    \item Dense Layer (2 units, softmax activation): For binary image classification, a final Dense Layer with 2 units and softmax activation is used, producing output probabilities for the two classes (real and fake).
\end{itemize}
\subsubsection{Text-Image Fusion Model}
Our proposed fusion Model comprises the following layers:
\begin{itemize}
    \item Input 1: Output of LSTM Text Model: The output of the LSTM Text Model serves as one of the model's inputs. It represents the textual features extracted from news headlines and processed through the text model's layers.
    \item Input 2: Output of VGG16 Image Model: The output of the VGG16-based Image Model serves as the second input. It embodies the image features extracted using deep learning techniques from the associated news article images.
    \item Concatenate Layer (combines text and image model outputs): The Concatenate Layer seamlessly combines the outputs from both the Text and Image Models. This fusion step ensures that both textual and visual information are taken into account for subsequent analysis.
    \item Dense Layer (64 units, ReLU activation): After fusion, a dense layer is introduced. This layer is instrumental in capturing intricate relationships and patterns that may arise from the fusion of features.
    \item Dense Layer (output: num classes, softmax activation): The final Dense Layer produces the model's output. The number of units in this layer matches the number of classes for classification. The softmax activation function is used to derive class probabilities.
\end{itemize}
\section{Experiments, Results, and Discussions}
This section explores the specifics of the implemented experiments of our proposed approach. All the experiments were conducted on NVIDIA A100 GPU compute using Python 3.9 with Tensorflow. The label encoding was done using the scikit-learn library, and the embeddings were generated using gensim and a local copy of VGG16. The number of epochs was set to 10 for this experiment, and the vector size to 300. The LSTM layers in the text-based model were configured with 128 units for the first layer, followed by 64 units for the second layer. The number of units in an LSTM layer affects the model's capacity to capture and learn from sequential data. A higher number will allow the model to capture more intricate patterns but can also lead to overfitting.
A dense layer with 64 units was added after the LSTM layers with a dropout rate of 0.5 was chosen. For the VGG-16 model, the dense layer had 256 units, another dense layer had 64 units, and an output layer with 2 units. The output layer consists of two units, one for real news and one for fake news, making it a binary classification task. The dropout was set to 0.5, and the Adam optimizer was chosen for the image model. Adam is an adaptive learning rate optimization algorithm that performs well in various deep learning tasks.
\begin{figure}[htb]
     \centering
     \begin{subfigure}[b]{0.45\textwidth}
         \centering
         \includegraphics[width=\textwidth]{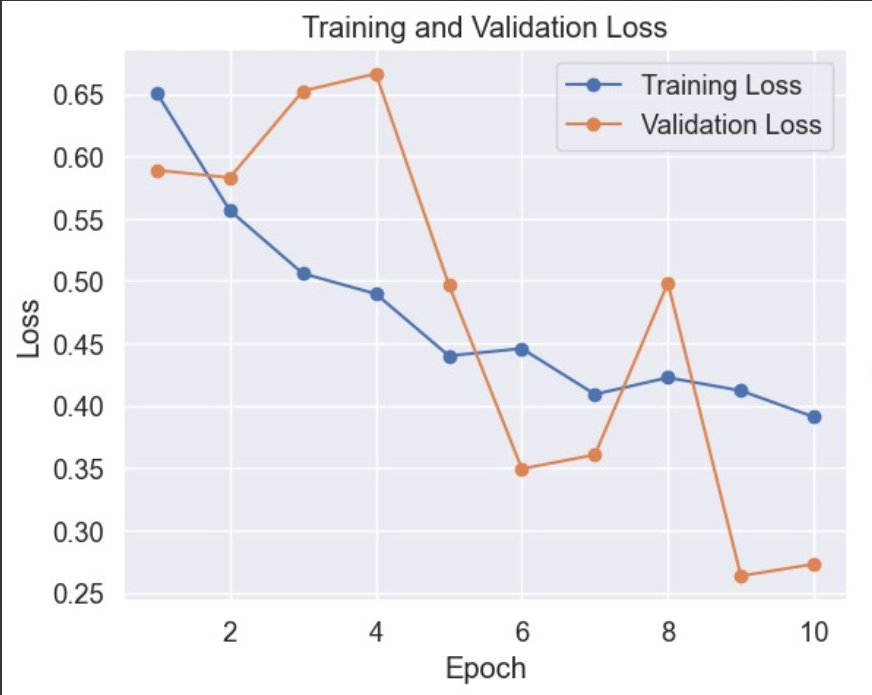}
         \caption{Graph showing training and validation loss}
     \end{subfigure}
     \hfill
     \begin{subfigure}[b]{0.45\textwidth}
         \centering
         \includegraphics[width=\textwidth]{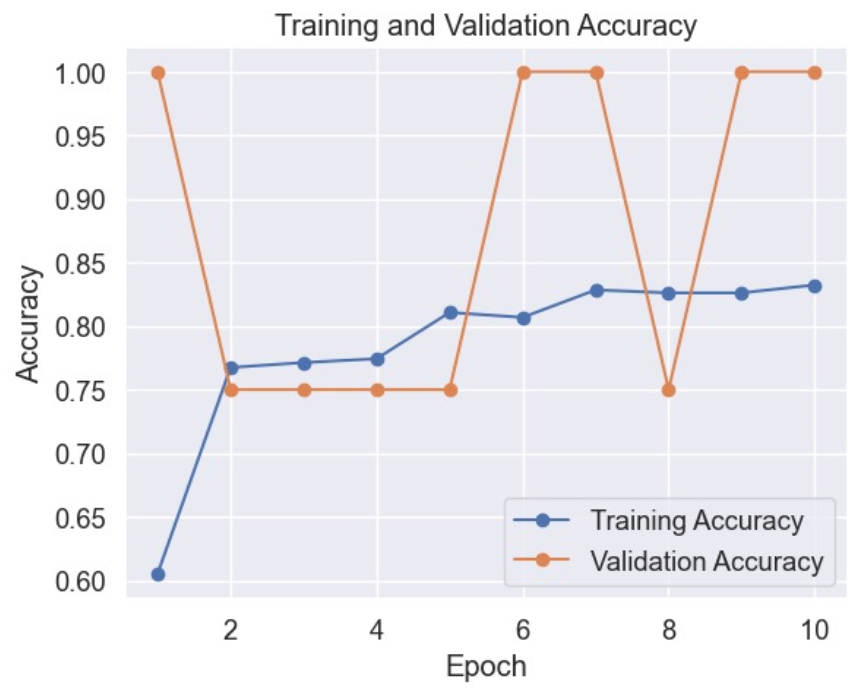}
         \caption{Graph showing training and validation accuracy}
     \end{subfigure}
        \caption{Graphs showing training loss, validation loss, training accuracy, and validation accuracy of the proposed approach}
\end{figure}
To create a prototype of the research output, we have also implemented a minimum working version and hosted the same using the Streamlit framework, which provided an intuitive and user-friendly interface for end-users. 
\begin{table}[]
\centering
	\caption{The precision, Recall, and F1 score produced by the proposed approach for different classes}
\begin{tabular}{|l|l|l|l|l|}
\hline
                      & \textbf{Precision} & \textbf{Recall} & \textbf{f1-score} & \textbf{support} \\ \hline
\textbf{Not fake}            & 0.70               & 0.62            & 0.66              & 284              \\ \hline
\textbf{Fake}            & 0.64               & 0.72            & 0.68              & 268              \\ \hline
\textbf{accuracy}     &                    &                 & 0.67              & 552              \\ \hline
\textbf{macro avg}    & 0.67               & 0.67            & 0.67              & 552              \\ \hline
\textbf{weighted avg} & 0.67               & 0.67            & 0.67              & 552              \\ \hline
\end{tabular}
\end{table}

\begin{figure}[htb]
	\centering
	\includegraphics[width=8cm,height=7cm]{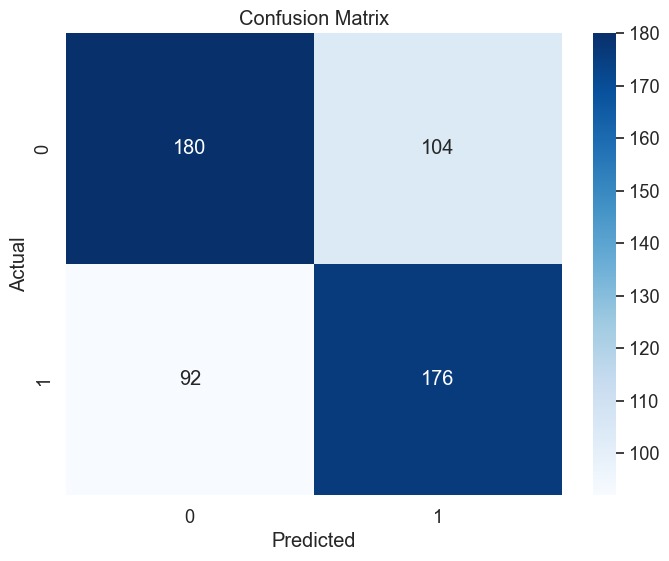}
	\caption{Confusion matrix for the classifier}
	\label{fig:fig5}
    \end{figure}
\section{Conclusion and Future Work}
This work proposed a multi-modal framework for identifying fake news in the Malayalam language. We have proposed and implemented a hybrid approach that combines textual and image features for identifying fake news from multi-modal sources. Our proposed approach showed reasonably good accuracy in the classification task comparable with any such baseline models. As the results are promising, the authors may continue working on improving the accuracy of the model by training the same with more data to cover a wide range of news from multiple domains. Incorporating explainability into the model will also be an interesting research dimension.
\bibliographystyle{unsrtnat}
\bibliography{references} 
\section*{Appendix}
\begin{figure}[htb]
	\centering
	\includegraphics[width=8cm,height=7cm]{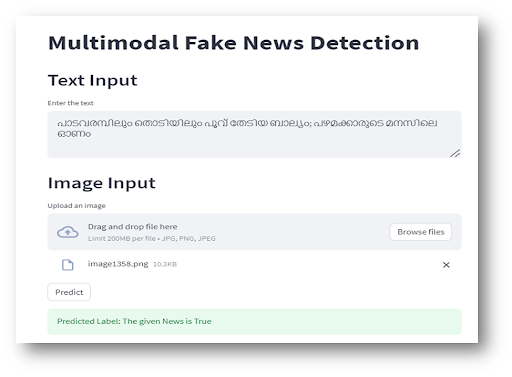}
	\caption{Output from our hosted streamlit application for a news which is not actually fake}
	\label{fig:fig6}
    \end{figure}
    \begin{figure}[htb]
	\centering
	\includegraphics[width=8cm,height=7cm]{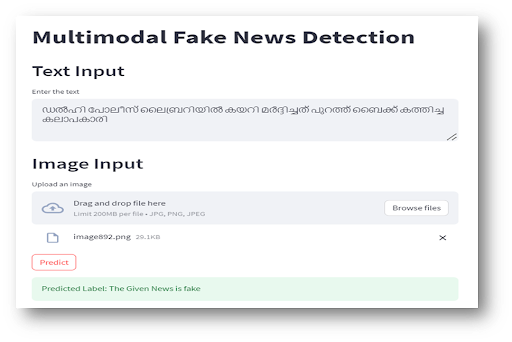}
	\caption{Output from our hosted streamlit application for a news which is fake}
	\label{fig:fig7}
    \end{figure}
\end{document}